\documentclass[sigconf]{acmart}

\AtBeginDocument{%
  }

\setcopyright{acmlicensed}
\copyrightyear{2018}
\acmYear{2018}
\acmDOI{XXXXXXX.XXXXXXX}

\acmConference[WWW '25]{Make sure to enter the correct
  conference title from your rights confirmation emai}{April 28--May 2,
  2025}{Sydney, Australia}
\acmISBN{978-1-4503-XXXX-X/18/06}

\usepackage{pifont}
\usepackage{hyperref}
\usepackage{caption}
\usepackage{xspace}
\usepackage{tcolorbox}
\usepackage{mathtools}
\usepackage{pifont}
\usepackage{xcolor}
\definecolor{mycolor1}{HTML}{edd8f3}
\definecolor{mycolor2}{HTML}{5571c0}
\definecolor{mycolor3}{HTML}{ccdfb5}

\newcommand{\datasetFont}{\text}
\newcommand{\ours}{\datasetFont{GE-Chat}\xspace}

\begin{document}

\title{GE-Chat: A Graph Enhanced RAG Framework for Evidential Response Generation of LLMs}

\author{$\text{Longchao Da}^1$, $\text{Parth Mitesh Shah}^1$, $\text{Kuan-Ru Liou}^1$, $\text{Jiaxing Zhang}^2$, $\text{Hua Wei}^{1*}$}

\email{{longchao, pshah113, kliou, hua.wei}@asu.edu,  tabzhangjx@gmail.com}
\affiliation{%
  \institution{$\text{Arizona State University}^1$, $\text{New Jersey Institute of Technology}^2$}
  \city{$\text{Arizona}^1$, $\text{New Jersey}^2$}
  \country{USA}
}





\renewcommand{\shortauthors}{Longchao et al.}

\begin{abstract}
 Large Language Models are now key assistants in human decision-making processes. However, a common note always seems to follow: "LLMs can make mistakes. Be careful with important info." This points to the reality that not all outputs from LLMs are dependable, and users must evaluate them manually. The challenge deepens as hallucinated responses, often presented with seemingly plausible explanations, create complications and raise trust issues among users. To tackle such issue, this paper proposes GE-Chat, a knowledge \underline{G}raph enhanced retrieval-augmented generation framework to provide \underline{E}vidence-based response generation. Specifically, when the user uploads a material document, a knowledge graph will be created, which helps construct a retrieval-augmented agent, enhancing the agent's responses with additional knowledge beyond its training corpus. Then we leverage Chain-of-Thought (CoT) logic generation, n-hop sub-graph searching, and entailment-based sentence generation to realize accurate evidence retrieval. We demonstrate that our method improves the existing models' performance in terms of identifying the exact evidence in a free-form context, providing a reliable way to examine the resources of LLM's conclusion and help with the judgment of the trustworthiness. The \textcolor{blue}{datasets} are released at~\footnote{\url{https://drive.google.com/drive/folders/1kNcsn1v0KH_srgL8w-NKvZM25o3onHBj?usp=sharing}}.
\end{abstract}



\maketitle

\section{Introduction}\label{sec:intro}
Large Language Models (LLMs) have demonstrated remarkable capabilities in multi-round conversational chats, including understanding questions~\cite{hu2024bliva}, generating responses, and performing reasoning or inference using the given context~\cite{yuan2024advancing}, \cite{zhang2024regexplainer}. The rapid advancement of LLMs has enabled a wide range of applications across various domains, including customer support~\cite{folstad2019chatbots, nandkumar2024enhancing}, virtual assistance~\cite{wei2024improving}, and augmented agents with tool-usage capabilities~\cite{da2024open}.

However, even though the LLMs are trained on massive expert corpus covering a wide range of topics, they are not immune to generating incorrect or misleading information, as called `hallucination'~\cite{ji2023towards}. It is always suggested to users on LLM interface that, `LLMs can make mistakes. Check important information carefully', which underscores the dilemma of seeking to benefit from LLMs' capabilities while contending with the challenge of the quality of the responses.

\begin{figure*}
    \centering
    \includegraphics[width=0.99\linewidth]{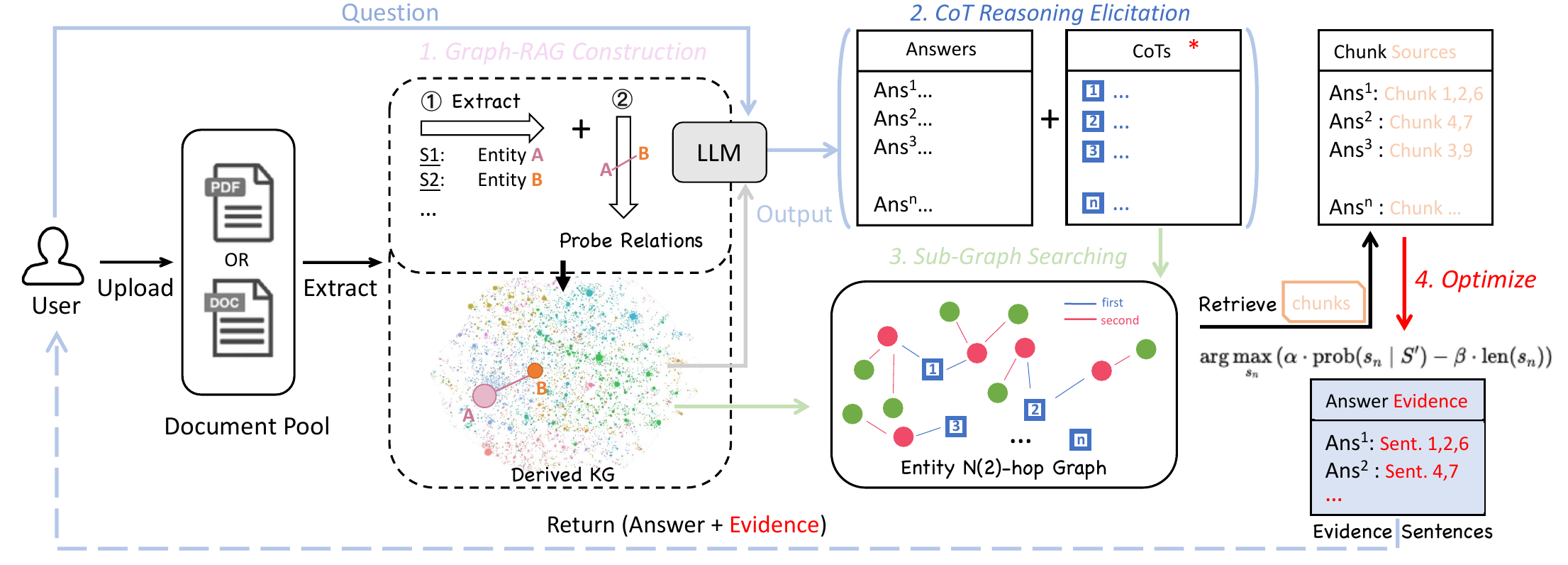}
    \caption{The overview of the \ours framework. As shown in this pipeline, when user uploads the document, it is used for the 
    \textcolor{mycolor1}{1. Graph-RAG Construction}, which contains two main steps using LLMs, \ding{192} Extract the entities A, B, etc., from the document chunks, then \ding{193} Probe the contextual relations between these entities. Then a derived Knowledge Graph is formed and used for question answering. In order to realize evidence generation on Graph-RAG, \textcolor{mycolor2}{2. CoT Reasoning Elicitation} is proposed to elicit the reasoning chain for answers. Then we have \textcolor{mycolor3}{3. Sub-Graph Searching} based on Entity Matching, and N-hop Relations Probing, this sub-graph contains entities and relations used to retrieve the \textcolor{orange}{source chunks}, guaranteeing originality of content. We then apply an \textcolor{red}{4. Optimize} objective to balance meaningfulness and conciseness to get the high-quality evidence.}
    \label{fig:method}
\end{figure*}

The solutions can be roughly divided into two aspects: One is to ground the LLMs' output to the factuality, either by fine-tunning~\cite{wang2024mitigating} or information retrieval augmentation~\cite{shuster2021retrieval}. Fine-tunning on the LLMs to reduce hallucination takes more resources and can be computationally expensive, meanwhile, it is not applicable to black-box commercial LLMs. And information-retrieval-based methods take more steps for multi-resource factuality checking and grounding, such as~\cite{ding2024retrieve, ji2023survey}. This will rely on the information source to vote for the trustworthiness of the generated responses and requires more query times from other APIs or Knowledge bases. The other direction is to understand the confidence/uncertainty of LLMs' responses, such as the training uncertainty prediction layers or applying post-hoc response level uncertainty quantification~\cite{da2024llm, lin2023generating}. These methods tend to quantify a measurement value for the LLMs, it provides a convenient way to compare the performance among LLMs, but not straightforward enough for users to decide whether to believe the LLMs' conclusion.

Regarding this shortcoming, some work started with evidence matching the generated responses. Essentially, when a user uploads a document and performs a question asking a certain LLM, it tends to highlight the corresponding raw context from a document that is best relevant to the response, thus guiding the user to understand where the conclusion is drawn from the original document. However, the current method, such as ~\cite{lin2024revolutionizing,han2024ragqaarenaevaluatingdomain} uses the direct LLM responded source of the evidence~\cite{saad2023pdftriage, da2024evidencechat}, when faced with a redundant response, it can only perform on the chunk-level resource highlight, which often gives a whole paragraph of relevant context without fine concentration. We also empirically observe that the performance of LLMs in reflecting their source evidence varies significantly depending on their instruction-following capability. However, smaller language models that excel at answering specific domain questions but lack instruction fine-tuning often struggle to effectively highlight relevant information for users to reference. 

To resolve the above issues, this paper proposes a framework named \ours that provides users with evidence of an LLM's response in a more accurate and generalizable way. Different from existing work, this method not only poses constraints on the derived source that it must come from the raw context, but also, \ours provides sentence-level fine-grained identification to accurately mark out the evidence supporting the LLMs' conclusion. What's more, this framework can be applied to any LLM with outstanding evidence retrieval ability (even a small model with limited fine-tuning on instruction-following ability, our framework still helps to highlight the evidence of conclusion). We compare with the direct LLM source reflection of the response, and our method essentially improves on the evidence retrieval performance. A \textcolor{blue}{demo video} and \textcolor{blue}{dataset}~\footnote{\href{https://drive.google.com/drive/folders/1kNcsn1v0KH_srgL8w-NKvZM25o3onHBj?usp=sharing}{Click the link to the demo video}} are provided for better understanding.

\section{Approach}
In this section, we will discuss the details of the  \textbf{\ours} framework. This work builds upon the graph-based retrieval augmented generation agent (Graph-RAG), and then provides a three-step paradigm for better evidential response generation, we will introduce the process of constructing the RAG agent first (in Sec.~\ref{sec:graphrag}), and then explain three components in our framework (Sec.~\ref{sec:cot} to  ~\ref{sec:optimize}).

\subsection{Graph-RAG construction}\label{sec:graphrag}
The Graph-RAG~\cite{larson2024graphrag} is unique in its ability to integrate external information through a structured knowledge graph, thus, supporting graph-based queries, and allowing for relational reasoning. Besides, Graph-RAG also does well in handling multi-hop reasoning, this feature can help us to find more than one-hop-related entities in the knowledge base, extract sub-graphs, and capture the relational semantics clusters. After a user uploads a document (.TXT/.DOC/.PDF) format, the metadata will be cut into the corpus chunks to temporarily store the file, then, we construct a knowledge graph $\mathcal{G}$ by two steps: \ding{192} extract entities from the chunks, and \ding{193} probe the relations among the entities. The LLM is used to achieve the knowledge graph, and the KG is used back to LLM as external information to make responses. We constructed a fast and lightweight Graph-RAG for a work basis following the implementations~\cite{guo2024lightrag}.

\subsection{CoT Reasoning Elicitation}\label{sec:cot}

\begin{tcolorbox}[colback=green!5,
                  colframe=black,
                  width=8.3cm,
                  arc=1mm, auto outer arc,
                  boxrule=0.05pt,
                  fontupper=\fontsize{9}{11}\selectfont, 
                 ]
\textcolor{red}{<<INST>><<SYS>>}

You are an agent to provide question answering tasks based on the provided document. 


\textcolor{red}{[Task]}

Your task is to generate answers to the user's question, please think step-by-step for the conclusion, and provide your thinking steps behind the output. 


\textcolor{red}{[Output Format]}

Answer: \{ [text] \}

Thoughts: \{1.[text] 2.[text] 3.[text] ... n.[text]\}

\end{tcolorbox}

This section introduces the Chain-of-Thought (CoT) reasoning inducer, which serves as a primary step in deriving the reasoning process. It is well acknowledged that the majority of the LLMs can automatically perform CoT~\cite{wei2022chain} to elicit the reasoning process, i.e., how they draw the conclusion step by step. We follow the same idea to induce the logic steps $ \textit{Logic\_steps} \{\textit{step}_1, \textit{step}_2, ..., \textit{step}_n\}$ from the LLM given a question $Q$ on a document. By designing a CoT template, we tend to achieve the following: 
\begin{equation}
    \textit{Answer}, \textit{Logic\_steps} = \text{CoT\_template}(\textit{Q}, \textit{Doc})
\end{equation}

The template is inspired by work~\cite{zhang2022automatic}, and is shown in the above green block. This step corresponds to the upper right part of Fig.~\ref{fig:method}, \texttt{CoT Reasoning Elicitation}, where each answer is associated with a CoT chain that explains the reasoning process step by step. These CoT chains, generated by RAG models, provide a logical structure to the responses but may not inherently align with the raw content of the submitted document. To ensure the evidence strictly originates from the provided source, a critical grounding step is introduced through sub-graph searching based on entity matching. This process anchors the CoT reasoning to specific entities and relationships within the knowledge graph, bridging the gap between generated content and its original context. By doing so, we enhance the trustworthiness and accuracy of the responses while maintaining a clear traceability to the original document.

\subsection{Efficient Sub-Graph Searching}\label{sec:subgraph}
\begin{figure}[h!]
    \centering
    \includegraphics[width=0.89\linewidth]{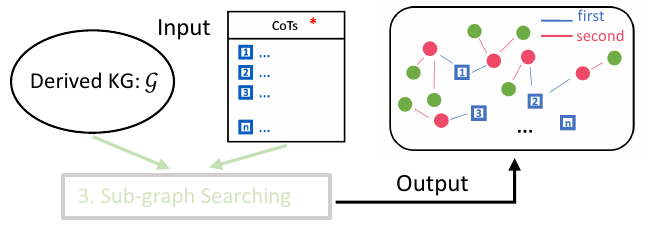}
    \caption{A Simple Abstraction for Sub-graph Searching.}
    \label{fig:subsearch}
\end{figure}

The sub-graph searching is conducted based on two resources: Derived KG $\mathcal{G}$ and CoTs as in Figure~\ref{fig:subsearch}. For each of the CoT results: $c_i \in \{c_1, c_2, c_3, ...., c_n\}$, the $\textcolor{blue}{c}_{\textcolor{blue}{i}}$ will be used to match the most relevant graph entities (Entity Matching), as shown in the output part, the blue boxes are the identical $c_i$, and the \textcolor{red}{\huge$\bullet$} is connected by blue edges, which is the first hop most relevant entities, this is what $c_i$ has been involved in the LLM's answer, then we relax this relation to further second-hop as shown in \textcolor{green}{\huge$\bullet$} green dot. This search is efficient for the CoT guidance and pre-calculated $\mathcal{G}$ for n-hop relationships probing (in contrast to the whole document-range global search). Finding this entity sub-graph is like finding an anchor that leads to the original source chunks, we can perform \texttt{\textcolor{orange}{Source Chunk Retrieval}} to get several chunks for each $c_i$ in CoTs. This step bridges the made-up CoT content with the source content of documents by finding the anchor entity. However, the chunk-level descriptions leave space for more fine-grained evidence sentences, which we employ an optimization objective to achieve in the following section. 

\subsection{Evidence Content Optimization}\label{sec:optimize}

As in Figure~\ref{fig:method}, based on the chunk sources (\textcolor{orange}{e.g., Chunk1,2,6} ), for each of the $c_i$ in CoTs, we expect to find a finer sentence from a certain chunk that best supports the answer sentence $S'$ with minimal redundant information. We first formalize this problem, and then inspired by the trade-off between meaningfulness and conciseness, we provide a solution in corresponds to the \textcolor{red}{4. Optimize} in the Figure~\ref{fig:method} leveraging the entailment probability. 

\paragraph{Problem Setting}

Given a chunk of text containing \( n \) sentences, \( S = \{s_1, s_2, \dots, s_n\} \), and a target sentence \( S' \) which represents the sentence in the answering content, the objective is to find the best sentence \( s_{\text{best}} \in S \) that:
1. Maximizes the entailment probability \( \text{prob}(s_n \mid S') \), which measures how strongly \( s_n \) entails \( S' \),
2. Minimizes the sentence length \( \text{len}(s_n) \), encouraging concise representations. 

In order to balance these two criteria, we define an objective function \( \mathcal{F}(s_n) \), which assigns a score to each sentence \( s_n \) based on its contained meaning and conciseness. The score for each sentence \( s_n \) is given by:
\begin{equation}
    \mathcal{F}(s_n) = \alpha \cdot \text{prob}(s_n |- S') - \beta \cdot \text{len}(s_n)
\end{equation}

where \( \alpha \) and \( \beta \) are set as 0.5 to control the weight of the entailment probability, and penalty for longer sentences, respectively. We want to measure how much the generated evidence means similarly to the answer, a rational way is to calculate the entailment probability $\text{prob}(s_n |- S')$. We achieve this by using NLI model~\footnote{off-the-shelf DeBERTa-large model}, which provides a three-element tuple by taking two text pairs $s_n$ and $S'$: $
    [\textit{logit}_{cont}, \textit{logit}_{neut}, \textit{logit}_{ent}] = \overrightarrow{\textit{NLI } }
 (s_{n}, S') $. 
The output is processed by transforming into the probability through: 
\begin{equation}
    \textbf{p} = \textit{Softmax} (\textit{logit}_{cont}, \textit{logit}_{neut}, \textit{logit}_{ent})
\end{equation}
then we can calculate the $\overrightarrow{p_{ent}} (s_{n}, S') = p (s_{n} \vdash S') = \textbf{p}_3$ as the entailment probability. The optimal sentence \( s_{\text{best}} \) is the one that maximizes \( \mathcal{F}(s_n) \):
$s_{\text{best}} = \arg\max_{s_n \in S} \left( \mathcal{F}(s_n) \right)$
Using this objective, we can find the best evidence that supports the answers in the LLM's responses, and this action is performed in a small chunk, which is not computationally expensive and can be deployed in real-time. 

The $s_{\text{best}}$ will be calculated for each of the answers, such as in Figure~\ref{fig:method}, the best evidence output for $\texttt{Ans}^1$ is a combination of sentences \texttt{Sent.1, 2, 6}. And this will be returned back to users together for users to understand which part of the answer comes with the evidence supported and which part lacks such trustworthy information, helping practitioners understand the reliability of generated content.

\textbf{Complexity analysis} The computational complexity of \ours involves three main steps: entity extraction, relation probing, and sub-graph searching. Entity extraction for $n$ chunks with an average length of $l$ words has a complexity of $O(n \cdot l)$. Relation probing for $m$ entities to construct a knowledge graph is $O(m^2)$. Sub-graph searching with $e$ edges for $k$-hop relations is $O(k \cdot e)$, where $k = 2$ in our implementation. The overall complexity is about $O(n \cdot l + m^2 + 2 \cdot e)$, dominated by $O(n \cdot l)$ and $O(m^2)$ in practical scenarios.

\begin{figure*}
    \centering
    \includegraphics[width=0.99\linewidth]{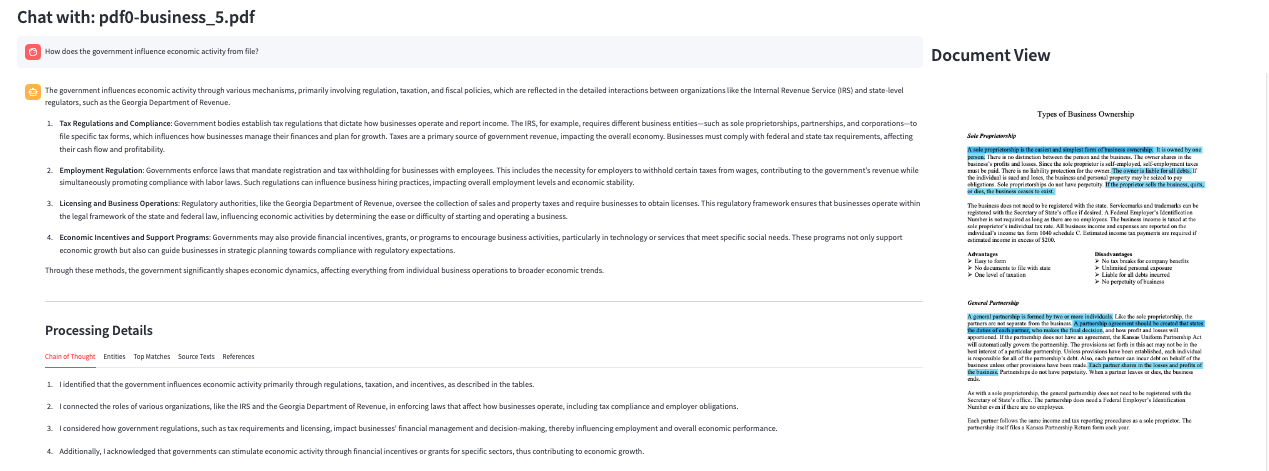}
    \caption{The demonstration of the deployed \ours framework.The user can upload PDF or relevant files, and the highlighted evidence comes along with the answers the LLMs made. For more examples please check the live demo video.}
    \label{fig:enter-label}
\end{figure*}

\section{Experiment}
\subsection{Experiment Setup}
\textbf{Dataset construction} To address the scarcity of evidence sources in prior research, we created a dataset with 1000 cases to evaluate evidence generation quality across 10 categories: Biology, Business, Chemistry, Computer Science, History, Management, Mathematics, Physics, Semiconductors, and Story. The dataset is structured along three dimensions: (1) PDF length—short (<10 pages), medium (10-100 pages), and long (>100 pages); (2) question types—Synthesis (integrating multiple parts), Structure (examining organization), and Term Explanation (defining specific concepts); and (3) human-annotated answers with corresponding evidence sentences, ensuring reliability and comprehensiveness. We tested our method on this dataset and have released \textcolor{blue}{dataset and videos}~\footnote{\href{https://drive.google.com/drive/folders/1kNcsn1v0KH_srgL8w-NKvZM25o3onHBj?usp=sharing}{Click the link to the testset and videos.}}   for public use with standard questions, groundtruth answers and evidence for reference.





\textbf{Evaluation Metric:} We refer to the existing work~\cite{eva1} that uses the cosine similarity to evaluate the relevance of the generated text (generated evidence) with the correct text (correct evidence), and use the conciseness score~\cite{eva2} to quantify the LLM's ability to find the corresponding evidence precisely.  In general, we have the following calculation that combines the two aspects of evaluation on $\text{Evidence}_\text{score}$: 
\begin{equation}
    \text{Evidence}_{\text{score}} = \frac{1}{N} \sum_{i=1}^{N} \left[ \cos(E_i, E_{gt_i}) \cdot \min\left(1, \frac{L_{gt_i}}{L_i}\right) \right]
\end{equation}
where the $E_i$ is the embedding of the generated evidence for question $i$, and $E_{gt_i}$ is the groundtruth evidence. The first term measures the cosine similarity of two shreds of evidence, the larger, the better, and $L_{gt_i}$ is the length of text for groundtruth evidence while the $L_{i}$ is the generated evidence, $\frac{L_{gt_i}}{L_i}$ measures how concise the generated evidence is given the relevance on meaning from cosine similarity.

\subsection{Experiment Result}
 
\begin{figure}[h!]
    \centering    \includegraphics[width=0.60\linewidth]{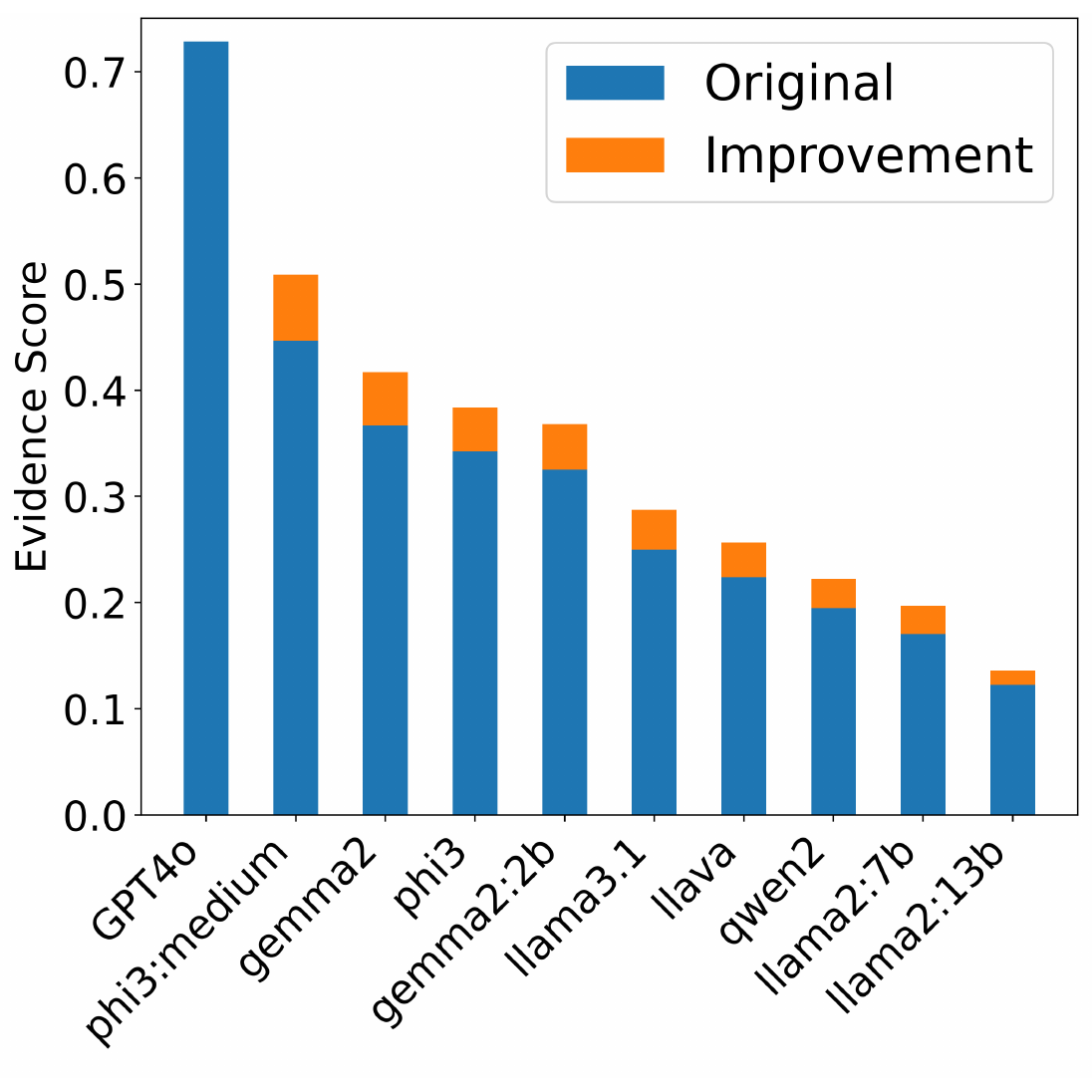}
    \caption{The comparison between the original LLMs evidence generation and LLMs with \ours framework.}
    \label{fig:bar}
\end{figure}

From the experiment, we observe that the direct evidence retrieval ability of GPT4o is the best, while other models perform much worse, especially because too many words are generated that prevent a concise evidence presentation. Then we conducted experiments for comparison, we applied our method \ours to existing the models except for GPT4o (this is because the GPT4o is involved and used in the ground-truth reference generation process, even though we added human-correction, we still aim to avoid potential bias or unfair advantages). The results are shown in Fig.~\ref{fig:bar}. It is visible that, compared to original models, applying the \ours framework could consistently improve the performance of each model's evidential-based responses.


\section{Conclusion}


In this paper, we presented \ours, a novel framework addressing the trustworthiness of LLMs by introducing a rigorous method for evidence retrieval and verification. Through hard constraints on source derivation and sentence-level highlight capabilities, \ours significantly enhances the reliability of LLM-generated responses. Our evaluation across ten diverse LLMs, both open and closed-source, demonstrates its robustness, versatility, and broad applicability. By offering a transparent and user-friendly approach, \ours contributes to making AI systems more reliable and trustworthy, paving the way for responsible deployment in critical decision-making processes.


\bibliographystyle{ACM-Reference-Format}
\bibliography{sample-base}

\end{document}